\documentclass{article}

    \PassOptionsToPackage{numbers, compress}{natbib}

    \usepackage[preprint]{neurips_2025}

\usepackage[utf8]{inputenc} 
\usepackage[T1]{fontenc}    
\usepackage{hyperref}       
\usepackage{url}            
\usepackage{booktabs}       
\usepackage{amsfonts}       
\usepackage{nicefrac}       
\usepackage{microtype}      
\usepackage{xcolor}         
\usepackage{amsmath}
\usepackage{amssymb}
\usepackage{multirow}
\usepackage{algorithm}
\usepackage{algpseudocode}
\usepackage{enumitem}
\usepackage{graphicx}
\usepackage{booktabs}

\usepackage{amsmath} 
\usepackage{amsfonts} 
\usepackage{amssymb}  

\title{DeepTraverse: A Depth-First Search Inspired Network for Algorithmic Visual Understanding}

\author{%
  Bin Guo \\
  Department of Electrical and Computer Engineering\\
  University of Texas at Dallas\\
  Richardson, TX 75080 \\
  \texttt{bin.guo93@utdallas.edu} \\
  \And
  John H.L. Hansen \\
  Department of Electrical and Computer Engineering\\
  University of Texas at Dallas\\
  Richardson, TX 75080 \\
  \texttt{john.hansen@utdallas.edu} \\
}

\begin{document}

\maketitle

\begin{abstract}
Conventional vision backbones, despite their success, often construct features through a largely uniform cascade of operations, offering limited explicit pathways for adaptive, iterative refinement. This raises a compelling question: can principles from classical search algorithms instill a more algorithmic, structured, and logical processing flow within these networks, leading to representations built through more interpretable, perhaps reasoning-like decision processes? We introduce DeepTraverse, a novel vision architecture directly inspired by algorithmic search strategies, enabling it to learn features through a process of systematic elucidation and adaptive refinement distinct from conventional approaches. DeepTraverse operationalizes this via two key synergistic components: recursive exploration modules that methodically deepen feature analysis along promising representational paths with parameter sharing for efficiency, and adaptive calibration modules that dynamically adjust feature salience based on evolving global context. The resulting algorithmic interplay allows DeepTraverse to intelligently construct and refine feature patterns. Comprehensive evaluations across a diverse suite of image classification benchmarks show that DeepTraverse achieves highly competitive classification accuracy and robust feature discrimination, often outperforming conventional models with similar or larger parameter counts. Our work demonstrates that integrating such algorithmic priors provides a principled and effective strategy for building more efficient, performant, and structured vision backbones.
\end{abstract}

\section{Introduction}
\label{sec:introduction}

The landscape of computer vision has been continually reshaped by the evolution of backbone architectures, from the foundational Convolutional Neural Networks (CNNs)~\cite{9010309,chollet2017xception,he2022tackling,li2021micronet,sifre2014rigid,singh2019hetconv} to the more recent rise of Vision Transformers (ViTs)~\cite{dosovitskiy2020image, touvron2021training}. Although these paradigms have achieved remarkable success by learning powerful hierarchical features, their predominant operational mode often involves layered, uniform, and feed-forward processing of information. This typically results in feature abstraction pathways that are learned implicitly and that potentially lack the explicit, structured, and iterative refinement characteristic of algorithmic problem solving. Consequently, a significant opportunity remains to explore architectures that embed more deliberate, step-by-step computational strategies, moving beyond conventional pattern recognition towards more profound algorithmic visual understanding.

\begin{figure}
  \centering
  \includegraphics[width=\textwidth]{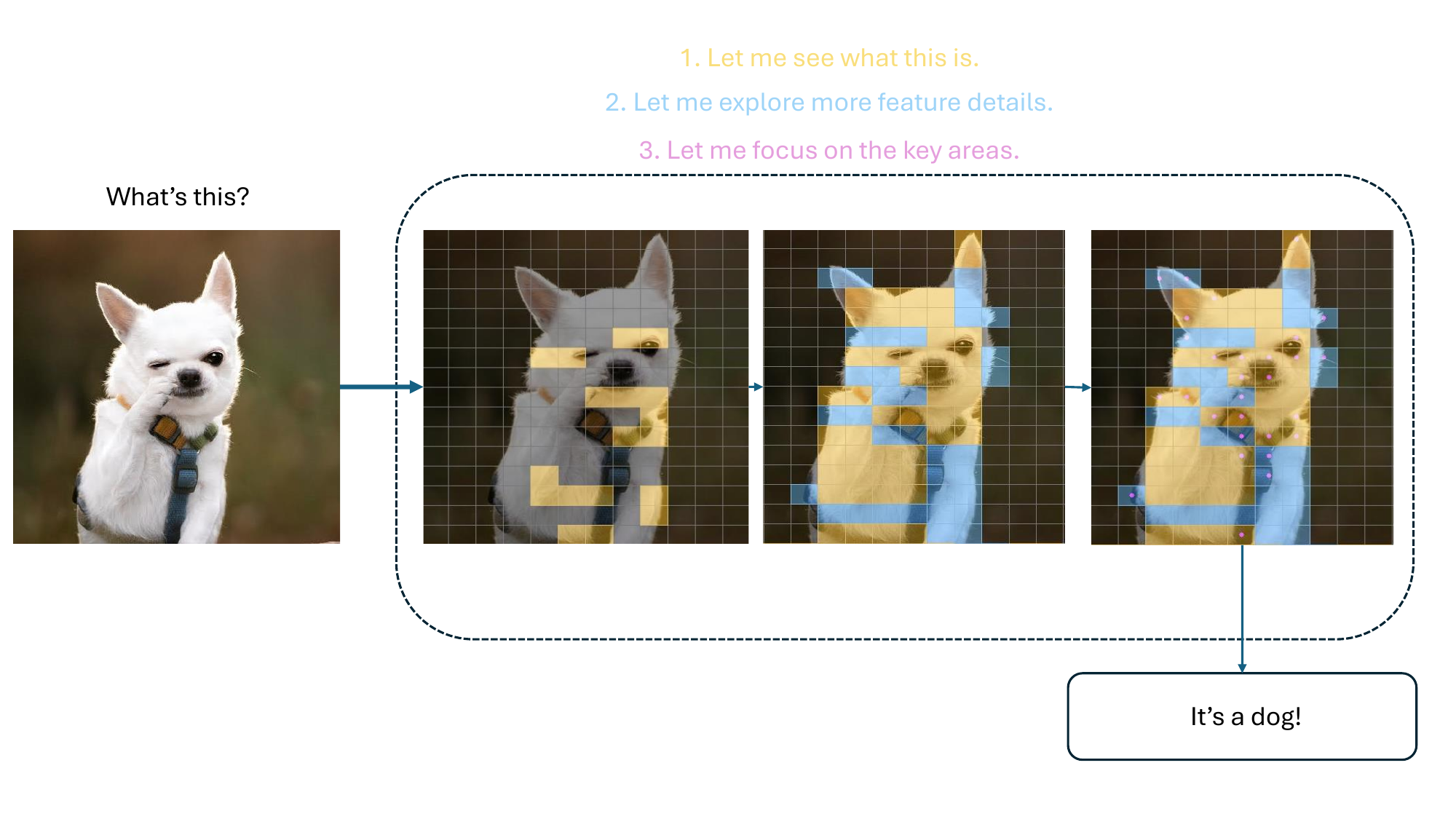} 
  \caption{DeepTraverse reasoning process: Illustrating how the architecture employs depth-first feature exploration and adaptive recalibration to process visual information in a systematic, algorithmic manner.}
  \label{fig:reasoning_process}
\end{figure}

This paper introduces DeepTraverse, a novel vision backbone that pioneers a new design philosophy: integrating classical algorithmic principles, specifically drawing core operational logic from Depth-First Search (DFS)~\cite{10326015}, into the heart of visual representation learning. We posit that by structuring network computations to emulate the systematic exploration and evaluative nature inherent in such algorithmic strategies, we can foster a more profound level of algorithmic visual understanding. Instead of relying solely on increasing depth or applying generic attention mechanisms, DeepTraverse is architected to actively and methodically navigate the feature space. (1) Recursive Exploration modules that perform systematic feature elucidation by iteratively delving into feature pathways with shared parameters, simulating a depth-first traversal to uncover intricate details and context along promising representational avenues; and (2) Dynamic Recalibration modules, inspired by the evaluative nature of DFS backtracking, which execute adaptive contextual recalibration by dynamically assessing and re-weighting the importance of explored feature channels based on progressively accumulated global and local evidence. Together, these components instantiate a paradigm of systematic, depth-first feature elucidation and adaptive contextual recalibration, forming the bedrock of DeepTraverse's approach. Figure~\ref{fig:reasoning_process} demonstrates the reasoning process of DeepTraverse.

The algorithmic interplay within DeepTraverse, driven by this paradigm of systematic feature elucidation and adaptive recalibration, facilitates a more dynamic and structured feature abstraction process. Recursive exploration allows efficient deep feature mining without a linear increase in unique parameters, while the recalibration mechanism enables the network to perform context-aware evaluation and refinement of the generated representations. This approach contrasts with conventional models, where feature salience is often determined by static weights or less targeted attention patterns. By explicitly modeling a principled, search-like process, DeepTraverse aims to construct feature hierarchies that are not only effective but also potentially more interpretable due to their structured, iterative formation.
Our extensive experiments on standard image classification benchmarks, including across a diverse set of datasets featuring varied image resolutions, scales, and complexities (such as CIFAR-10, CIFAR-100, ImageNet-64, and ImageNet-1k), demonstrate the efficacy of DeepTraverse. It consistently achieves a superior performance-to-parameter trade-off, outperforming various established CNN and lightweight architectures, thereby validating the benefits of integrating such algorithmic search principles into vision backbone design.

The main contributions of this work are:
\begin{itemize}[leftmargin=*]
    \item We introduce DeepTraverse, a novel vision-backbone paradigm. Inspired by classical algorithmic strategies, particularly Depth-First Search, it embeds an explicit, multi-step process of principled feature elucidation and adaptive recalibration, transforming visual feature extraction into an active, iterative exploration and refinement journey.
    \item We design and implement unique architectural components: recursive exploration modules for methodical feature elucidation and dynamic recalibration pathways for adaptive evidence refinement that enable the network to systematically trace, evaluate and adaptively construct feature representations, fostering a more structured, dynamic, and potentially interpretable mode of visual information processing.
    \item We empirically validated our proposed approach through extensive experiments on diverse computer vision tasks. Specifically, DeepTraverse, a new family of vision backbones introduced herein, demonstrates a favorable performance-computation trade-off compared to state-of-the-art methods across multiple datasets, including ImageNet-1K under a supervised training regime.
\end{itemize}

\section{Related Works}
\label{sec:related_works}

\textbf{Vision Backbone CNNs.} The pursuit of efficient and potent Convolutional Neural Networks (CNNs)~\cite{9010309,chollet2017xception,he2022tackling,li2021micronet,sifre2014rigid,singh2019hetconv,zhang2017interleaved,zhuo2022semi} remains central, especially for resource-aware applications. CNN efficiency is commonly enhanced by optimizing convolutional operators, such as group convolution~\cite{krizhevsky2012imagenet} and depthwise separable convolutions~\cite{sifre2014rigid}. These innovations underpin influential mobile-oriented frameworks like MobileNets~\cite{howard2019searching,howard2017mobilenets,sandler2018mobilenetv2}, ShuffleNets~\cite{ma2018shufflenet,zhang2018shufflenet}, GhostNet~\cite{han2020ghostnet}, and others~\cite{7780459, huang2018condensenet, tan2021efficientnetv2, chen2023run_cvpr, Ma_2024_CVPR}. While these refinements reduce parameters and FLOPs, DeepTraverse introduces a distinct paradigm: an algorithmically inspired information processing flow. Its efficiency and performance stem from prioritizing structured, iterative feature exploration via parameter-shared recursive modules and dynamic, context-aware recalibration, diverging from optimizing atomic computations towards a holistic, algorithmically guided visual understanding.

\textbf{Vision Transformers (ViTs).} The application of Transformer principles~\cite{vaswani2017attention,wen2022social} to vision, via Vision Transformers (ViTs)~\cite{dosovitskiy2020image}, marked a shift by powerfully modeling global image context. This spurred research into ViT efficacy, including refined training~\cite{steiner2021train,touvron2021training,touvron2022deit} and architectural innovations~\cite{graham2021levit,liu2022swin,liu2021swin,wang2021pyramid,zhong2022tree}. Much work addresses computational scaling and data requirements, leading to efficient variants~\cite{ali2021xcit,huang2022lightvit,lu2022soft,tang2022quadtree,vaswani2021scaling} and convolutional element integration~\cite{chen2022mobileformer,dai2021coatnet,srinivas2021bottleneck,cai2022efficientvit,li2022efficientformer,mehta2022separable,pan2022edgevits}. MLP-like structures~\cite{chen2021cyclemlp,lian2021asmlp,tolstikhin2021mlp}, sometimes with CNN-like biases~\cite{liu2022ready}, are also explored. Balancing global interaction with efficiency and data dependency remains debated~\cite{liu2022convnet,wang2022shift,mehta2021mobilevit}.

DeepTraverse offers a distinct approach to visual understanding and dependency modeling. Instead of Transformer-based dense global interactions, its DFS-inspired algorithmic structure—iteratively exploring feature pathways for contextual cues and adaptively recalibrating salience—presents a different information processing paradigm. This structured search and refinement diverges from pure ViTs and common hybrids, offering a novel way to integrate diverse features and balance understanding with efficiency.

\textbf{Algorithm-Inspired Neural Design.} Embedding algorithmic principles within neural networks can yield more structured, interpretable, and efficient models~\cite{10326015,battaglia2018relational, gregor2010learning, velivckovic2019neural}. DeepTraverse significantly advances this by translating Depth-First Search (DFS) strategies into learnable, dynamic vision backbone components. Its recursive modules emulate DFS's depth-first traversal for methodical, parameter-efficient feature analysis, while dynamic recalibration modules perform context-aware adjustments inspired by DFS backtracking. This principled, algorithmic feature abstraction moves beyond conventional scaling or generic attention~\cite{hudson2018compositional}. While DeepTraverse's iterative processing shares conceptual parallels with multi-step reasoning in language models like Chain-of-Thought~\cite{wei2022chain} and Tree of Thoughts~\cite{yao2023tree}, our core contribution is operationalizing DFS principles within a vision backbone for robust, efficient visual understanding. This offers a novel path to interpretable visual understanding~\cite{chen2019this} through an explicitly structured, adaptive computational process.

\section{Methodology}
\label{sec:methodology}

This section first formalizes the image classification task, then meticulously details the operational principles and constitution of DeepTraverse's key components: the DFS-inspired Exploration Block (DFS-EB) for feature pathway elucidation, the DFS-inspired Backtrack Block (DFS-BB) for adaptive recalibration, and the comprehensive DFSBlock that synergistically integrates these mechanisms.

\subsection{Problem Formulation}
\label{ssec:problem_formulation}

Image classification aims to categorize visual inputs into predefined classes. We formalize a classifier as a function $f : \mathbb{R}^{C \times H \times W} \rightarrow \mathbb{R}^{c}$, which maps an input image $x$ with $C$ channels, height $H$, and width $W$ to a $c$-dimensional output vector, where $c$ is the number of classes. Each element $f_l(x)$ represents the score or probability that image $x$ belongs to class $l$. The final prediction is determined by $\text{argmax}_l f_l(x)$. The image classification task is typically formulated as an optimization problem:
\begin{equation}
\min_{\theta} \mathcal{L}(f_{\theta}(x), y)
\label{eq:problem_formulation}
\end{equation}
where $\theta$ denotes the model parameters, $y$ is the ground truth label, and $\mathcal{L}$ is a loss function (e.g., cross-entropy). The objective is to learn parameters $\theta$ such that the model accurately assigns input images to their correct classes.

\subsection{Recursive Path Exploration Block}
\label{ssec:dfs_eb}

The DFS-Inspired Exploration Block (DFS-EB) forms the primary mechanism for feature pathway elucidation in DeepTraverse. Drawing conceptual parallels with the exploratory nature of Depth-First Search algorithms, this block iteratively refines feature representations by systematically delving deeper into the feature space. Unlike a literal graph traversal, DFS-EB operationalizes this exploration through a recursive residual learning paradigm, efficiently capturing representations at varying levels of abstraction and complexity.

Given an input feature map $X \in \mathbb{R}^{C_{in} \times H \times W}$, the DFS-EB first processes it through an initial feature extraction layer, $\Phi_{\text{extract}}$, to establish a foundational representation $F_0$:
\begin{equation}
F_0 = \Phi_{\text{extract}}(X)
\label{eq:phi_extract_application}
\end{equation}
This layer, comprising operations such as depthwise convolution ($K_d$), Batch Normalization ($B$), ReLU activation ($\delta$), optional Dropout ($D$), and pointwise convolution ($P$), transforms $X$ into $F_0 \in \mathbb{R}^{C_{out} \times H' \times W'}$:
\begin{equation}
\Phi_{\text{extract}}(X) = B(P(D(\delta(B(K_d(X))))))
\label{eq:phi_extract_definition}
\end{equation}
$F_0$ serves as the starting point for subsequent depth-first iterative refinement, analogous to selecting an initial node or path in a search algorithm. This transformation establishes the foundation upon which successive refinements will build, setting the stage for the subsequent recursive exploration process.

The essence of DFS-EB's feature elucidation capability resides in its recursive refinement process. The representation $F_0$ undergoes $R$ stages of iterative deepening, governed by a shared recursive transformation module $\Phi_{\text{recursive}}$. At each stage $i \in [1, R]$, the feature map is updated via a residual connection:
\begin{equation}
F_i = F_{i-1} + \Phi_{\text{recursive}}(F_{i-1})
\label{eq:recursive_step}
\end{equation}
Here, $\Phi_{\text{recursive}}(F_{i-1})$ represents the newly elucidated feature details or refined insights gained by further exploring the current representational pathway, akin to DFS advancing one step deeper along a search branch. The recursive transformation $\Phi_{\text{recursive}}$ typically shares a similar structure with $\Phi_{\text{extract}}$ (e.g., $B(P(\delta(B(K_d(F_{i-1})))))$) but operates on $F_{i-1}$ and critically reuses its parameters across all $R$ iterations. 

This parameter sharing is key to DeepTraverse's efficiency, allowing for an effective depth of computation ($R$ refinement steps) without a linear increase in unique parameters. The computational advantage becomes particularly significant as $R$ increases, enabling the network to perform deeper feature exploration with a constrained parameter budget. The additive integration ensures that knowledge from previous depths $F_{i-1}$ is preserved while being augmented by new information, mirroring how DFS systematically accumulates information while traversing deeper into a search space.

\begin{figure}
  \centering
  \includegraphics[width=\textwidth]{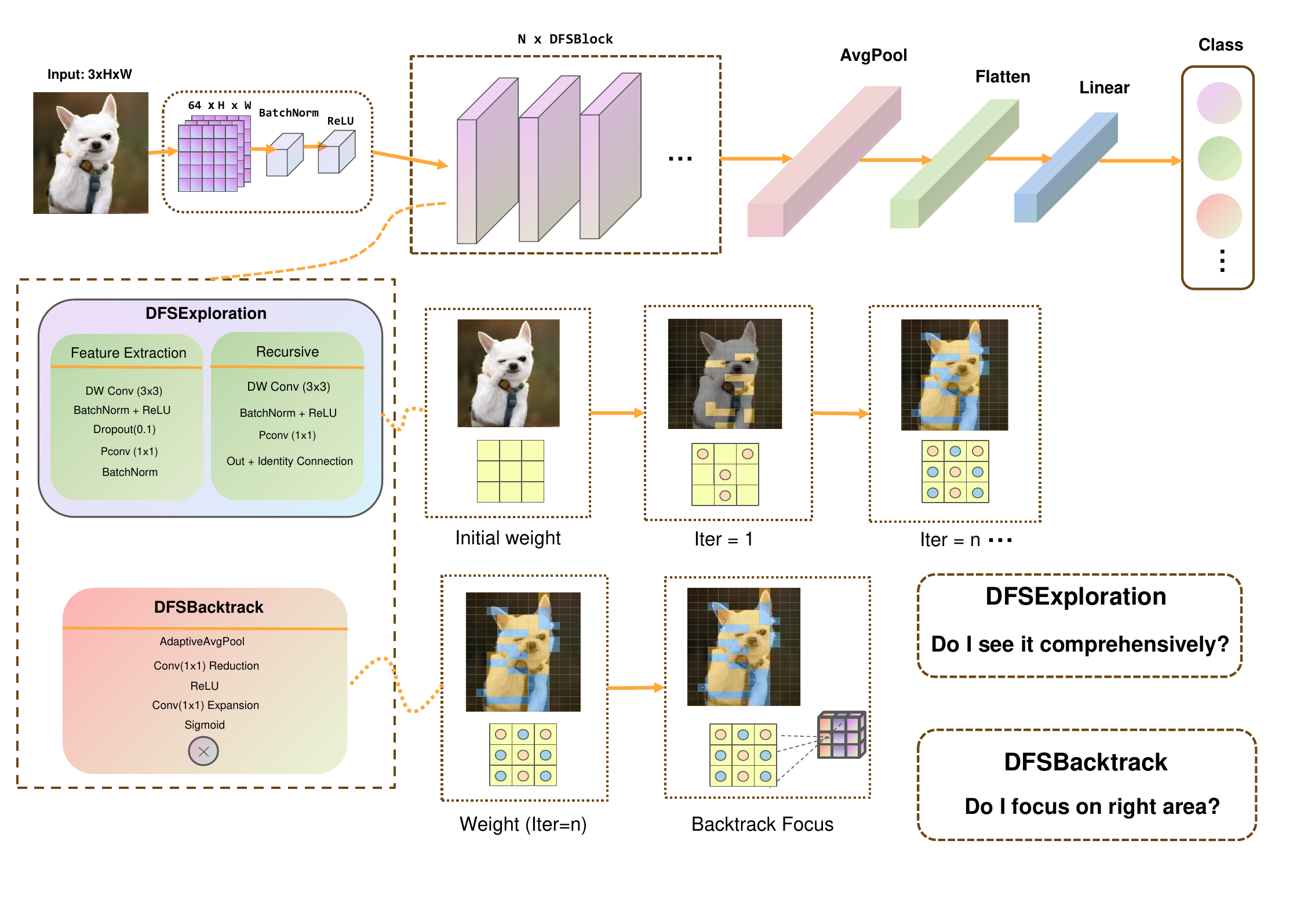}
  \caption{Structure of DeepTraverse, illustrating the integration of DFS-EB and DFS-BB modules within the overall architecture. The recursive pathways within DFS-EB and the contextual recalibration mechanism of DFS-BB together form the foundation of our algorithm-inspired approach to visual feature processing.}
  \label{fig:architecture}
\end{figure}

The final output of the DFS-EB, $Y_{EB} = F_R$, encapsulates the cumulative knowledge from this multi-stage exploration:
\begin{equation}
Y_{EB} = F_R = \Phi_{\text{extract}}(X) + \sum_{j=1}^{R} \Phi_{\text{recursive}}(F_{j-1})
\label{eq:dfs_eb_output_expanded}
\end{equation}
This iterative refinement, driven by a consistent (parameter-shared) strategy $\Phi_{\text{recursive}}$, provides an implicit ensemble of features from multiple effective depths~\cite{diskin2022performancedepthsearchalgorithm}. The progressive nature of this feature accumulation allows the network to capture patterns at multiple levels of abstraction within a single block, enhancing its representational capacity. The use of depthwise separable convolutions within $\Phi_{\text{extract}}$ and $\Phi_{\text{recursive}}$ further enhances computational efficiency, making deeper, more methodical recursive exploration feasible even within resource-constrained environments. This architectural choice enables DeepTraverse to maintain competitive performance while significantly reducing computational overhead compared to traditional approaches. Figure~\ref{fig:architecture} shows the structure of DeepTraverse.

\subsection{Adaptive Backtrack Module}
\label{ssec:dfs_bb}

Following the feature elucidation phase by DFS-EB, the DFS-Inspired Backtrack Block (DFS-BB) implements an adaptive contextual recalibration mechanism. This stage is conceptually analogous to the backtracking phase in DFS, where, upon reaching a certain depth or evaluative checkpoint, knowledge gathered along the explored path is used to reassess priorities and inform subsequent decisions. In DeepTraverse, DFS-BB leverages global contextual information derived from the explored features to adaptively re-weight feature channels, thereby emphasizing informative pathways and suppressing less salient ones.

Given an input feature map $F \in \mathbb{R}^{C \times H \times W}$ (which is typically the output $Y_{EB}$ from a preceding DFS-EB), DFS-BB first distills global spatial information into a channel descriptor $z$ using adaptive average pooling:
\begin{equation}
z = \text{AdaptiveAvgPool2d}(F) \in \mathbb{R}^{C \times 1 \times 1}
\label{eq:dfs_bb_squeeze}
\end{equation}
This descriptor $z$ acts as a condensed summary of the current state of feature exploration, capturing the global context that will inform the subsequent recalibration process. The spatial averaging operation effectively transforms the detailed spatial map into a channel-wise signature, providing a holistic view of each channel's overall activation patterns across the entire spatial domain.

This global descriptor $z$ is then transformed by a two-layer fully connected network (implemented as 1x1 convolutions with a bottleneck structure having reduction ratio $r$) to compute a channel-wise attention vector $s$:
\begin{equation}
s = \sigma(W_2 \delta(W_1 z)) \in \mathbb{R}^{C \times 1 \times 1}
\label{eq:dfs_bb_excitation}
\end{equation}
where $W_1 \in \mathbb{R}^{(C/r) \times C}$ and $W_2 \in \mathbb{R}^{C \times (C/r)}$ are learnable parameters, $\delta$ is ReLU, and $\sigma$ is the sigmoid activation. The dimensionality reduction and subsequent expansion through $W_1$ and $W_2$ enable the network to learn complex channel interdependencies while maintaining computational efficiency. The bottleneck ratio $r$ serves as a hyperparameter controlling the trade-off between model capacity and computational overhead. This attention vector $s$ represents the adaptive importance assigned to each feature channel based on the global "understanding" (captured by $z$) of the explored features.

Finally, the input feature map $F$ is recalibrated by performing an element-wise multiplication with $s$:
\begin{equation}
F' = F \odot s
\label{eq:dfs_bb_recalibration}
\end{equation}
This operation adaptively modulates the response of each feature channel in $F$ based on its learned global importance $s$. By applying the same scaling factor across all spatial locations within each channel, the recalibration preserves the spatial structure of features while enhancing or suppressing entire channels based on their contextual relevance. The DFS-BB thus enables the network to perform a form of computational reflection or self-evaluation on the features elucidated by the DFS-EB. 

By dynamically emphasizing channels that, in the global context of the explored representation, are deemed more relevant, and attenuating others, the DFS-BB refines the feature map, making it more discriminative for the downstream task. This process is analogous to how an algorithmic search might prune less promising branches or reinforce evidence from fruitful paths after a backtracking or evaluation step. The adaptive nature of this recalibration allows the network to allocate its representational capacity more efficiently, focusing computational resources on the most informative feature dimensions while reducing the influence of less relevant ones.

\subsection{Integrated DFS Block}
\label{ssec:dfs_block}

The DFSBlock serves as DeepTraverse's fundamental operational unit. It synergistically integrates the DFS-EB's principled feature elucidation with the DFS-BB's adaptive contextual recalibration, and is stabilized by a residual shortcut. Functioning as a complete algorithm-inspired micro-cycle, this block transforms an input $X_{in}$ to an output $X_{out}$ by orchestrating systematic exploration with context-aware refinement through a sequence of complementary computational stages.

Initially, the DFSBlock undertakes the Principled Exploration phase, where the DFS Exploration Block systematically investigates feature pathways through its recursive refinement mechanism. This stage transforms the input $X_{in}$ into a richly elucidated representation:
\begin{equation}
F_{\text{elucidated}} = \text{DFS-EB}(X_{in}; R, C_{out}, \text{stride})
\label{eq:dfsblock_exploration_detailed}
\end{equation}
The DFS-EB is configured with a specific number of recursive iterations $R$, output channels $C_{out}$, and potentially a stride parameter for spatial downsampling. The recursive nature of this exploration enables the block to build increasingly abstract representations through iterative refinement, effectively deepening the network without a proportional increase in parameters. This computational efficiency is particularly valuable in resource-constrained scenarios, allowing DeepTraverse to achieve competitive performance with a significantly reduced parameter footprint.

Following exploration, the Adaptive Contextual Recalibration phase employs the DFS Backtrack Block to evaluate the global context of the elucidated features and selectively emphasize the most informative channels:
\begin{equation}
F_{\text{recalibrated}} = \text{DFS-BB}(F_{\text{elucidated}})
\label{eq:dfsblock_recalibration_detailed}
\end{equation}
This stage performs a form of feature curation, dynamically adjusting the importance of different feature dimensions based on their contextual relevance. By enhancing channels that contain discriminative information and suppressing less informative ones, the DFS-BB increases the signal-to-noise ratio in the representation, making it more conducive to accurate downstream processing. The recalibration mechanism draws conceptual inspiration from the evaluation and prioritization process that occurs during the backtracking in the algorithmic search, where the path forward is determined based on insights gained from previous exploration.

To facilitate robust training of deep networks and preserve valuable information from earlier processing stages, the DFSBlock incorporates a Residual Integration mechanism via a shortcut connection. The original input $X_{in}$ is combined with the processed features $F_{\text{recalibrated}}$, creating a direct pathway for gradient flow during backpropagation. If the dimensions of $X_{in}$ do not match those of $F_{\text{recalibrated}}$ (due to changes in channels or spatial size by DFS-EB's $\Phi_{\text{extract}}$), a projection shortcut $S(X_{in})$ is applied to ensure dimensional compatibility:
\begin{equation}
S(X_{in}) = \begin{cases} \text{Proj}(X_{in}) & \text{if dimensions of } X_{in} \text{ and } F_{\text{recalibrated}} \text{ mismatch} \\ X_{in} & \text{otherwise} \end{cases}
\label{eq:dfsblock_shortcut_definition_detailed}
\end{equation}
The projection $\text{Proj}(X_{in})$ is typically implemented using a 1x1 convolution followed by Batch Normalization to adjust channels and, if necessary, an appropriate pooling or strided convolution for spatial alignment. This residual connection plays a crucial role in mitigating the vanishing gradient problem, especially in deeper configurations of DeepTraverse.

The final output of the DFSBlock, $X_{out}$, emerges from the synthesis of these complementary processing stages:
\begin{equation}
X_{out} = \delta(F_{\text{recalibrated}} + S(X_{in}))
\label{eq:dfsblock_final_output_detailed}
\end{equation}
where $\delta$ represents a non-linear activation function. This output encapsulates both the detailed feature insights from the exploration phase and the contextually weighted information from the recalibration phase, all anchored by the foundational information preserved through the shortcut connection.

By stacking multiple DFSBlocks in a hierarchical fashion, the DeepTraverse network constructs a sophisticated processing pipeline where this principled iterative approach to feature analysis operates at increasing levels of semantic abstraction. The architecture inherently prioritizes parameter efficiency through the extensive parameter sharing in the recursive exploration phase of DFS-EB and the common use of depthwise separable convolutions. Moreover, adaptive channel recalibration in DFS-BB allows the network to intelligently allocate its representational capacity, focusing on the most important information pathways. This structured and algorithm-inspired methodology enhances both the representational power and the potential for interpretability in deep neural networks for visual understanding, offering a compelling alternative to traditional CNN architectures and attention-based mechanisms for efficient feature learning.

\section{Experiments}
\label{sec:experiments}

\textbf{Baseline methods.} We compare DeepTraverse with several recent backbones on a series of computer vision tasks. These include ResNet20~\cite{7780459}, ResNet50~\cite{7780459}, GhostNet~\cite{han2020ghostnet}, DenseNet~\cite{huang2018condensenet}, EfficientNet~\cite{tan2021efficientnetv2}, MobileNetv3~\cite{howard2019searching}, ShuffleNetv2~\cite{ma2018shufflenet}, FasterNet~\cite{chen2023run_cvpr}, StarNet~\cite{Ma_2024_CVPR}.

\textbf{Datasets and Experiment setting.} To comprehensively evaluate the proposed method, multiple widely-used datasets are used including CIFAR-10 and CIFAR-100 with resolution of 32$\times$32, Tiny ImageNet with resolution of 64$\times$64, and a subset of ImageNet-1k. Following existing works, we train the model for image classification from scratch using the training recipe provided by previous work on the same datasets to ensure fair comparison. The computational platform is an NVIDIA RTX 2080 Ti. The training runs for 100 epochs. The initial learning rate is 0.1. We evaluate the models for classification accuracy and benchmark their size and FLOPs with scripts provided by the timm and thop libraries on the same hardware. More experimental settings and ablation studies are in supplemental materials.

\subsection{Results on CIFAR-100 and CIFAR-10}

Table \ref{tab:combined_results_vlines} demonstrates the superior performance of our DeepTraverse model across various computational scales. Under standard supervised training conditions, DeepTraverse consistently outperforms existing state-of-the-art models while maintaining exceptional efficiency. The lightweight version achieves 73.84\% accuracy on CIFAR-100 with only 0.26M parameters and 0.03G FLOPs, surpassing comparable models by a significant margin.
Most notably, DeepTraverse shows remarkable scaling properties, with our wide version reaching an impressive 82.20\% accuracy on CIFAR-100, representing a 1.39\% improvement over WideResNet while using only 39\% of its parameters (14.26M vs. 36.54M) and 34\% of its computational cost (1.78G vs. 5.25G FLOPs). This demonstrates the effectiveness of our architectural design at both small and large scales.
When compared to recent advanced architectures like StarNet and FasterNet, DeepTraverse establishes a new Pareto frontier in the accuracy-efficiency trade-off. Our model achieves 93.25\% accuracy on CIFAR-10, outperforming DenseNet (92.75\%) and StarNet (92.13\%) while requiring significantly fewer resources. These results validate our core hypothesis that integrating dynamic feature selection mechanisms within a traversal-inspired architecture enables significantly more effective information processing than conventional designs, establishing DeepTraverse as a breakthrough approach for resource-constrained image classification tasks.

\begin{table*}[t]
  \caption{Performance comparison across CIFAR-100 and CIFAR-10 datasets. Results show model efficiency metrics and Top-1 accuracy, with separate columns for dataset-specific parameters and FLOPs. Wide versions of models are included for CIFAR-100 only. Best results for each dataset are highlighted in bold.}
  \label{tab:combined_results_vlines}
  \centering
  \small 
  \newlength{\originaltabcolsepV}
  \setlength{\originaltabcolsepV}{\tabcolsep}
  \setlength{\tabcolsep}{2.2pt} 
  \begin{tabular}{lcccccc} 
    \toprule
    \multirow{2}{*}{Method} & \multicolumn{2}{c}{CIFAR-100} & \multicolumn{2}{c}{CIFAR-10} & \multicolumn{2}{c}{Top-1 Accuracy (\%)} \\
    \cmidrule(lr){2-3} \cmidrule(lr){4-5} \cmidrule(lr){6-7} 
    & FLOPs (G) & \# Param. (M) & FLOPs (G) & \# Param. (M) & CIFAR-100 & CIFAR-10 \\
    \midrule
    EfficientNet~\cite{tan2021efficientnetv2} & 0.12 & 4.14  & 0.34 & 0.26 & 73.14 & 90.20 \\
    ResNet20~\cite{7780459}       & 0.08 & 0.28  & 0.08 & 0.27 & 69.53 & 92.36 \\
    MobileNetV3~\cite{howard2019searching}    & 0.01 & 0.48  & 0.14 & 0.98 & 63.99 & 91.81 \\
    ShuffleNetV2~\cite{ma2018shufflenet}    & 0.09 & 1.35  & 0.09 & 1.26 & 71.83 & 91.49 \\
    DenseNet~\cite{huang2018condensenet}       & 0.20 & 0.60  & 0.20 & 0.60 & 73.02 & 92.75 \\
    GhostNet~\cite{han2020ghostnet}       & 0.04 & 2.76  & 0.04 & 0.28 & 72.46 & 91.79 \\
    StarNet~\cite{Ma_2024_CVPR}       & 0.28 & 2.70  & 0.28 & 2.68 & 72.27 & 92.13 \\
    FasterNet~\cite{chen2023run_cvpr}      & 0.81 & 0.55  & 0.81 & 0.54 & 63.91 & 89.61 \\
    DeepTraverse (Ours)             & 0.03 & 0.26  & 0.03 & 0.26 & \textbf{73.84} & \textbf{93.25} \\
    \midrule 
    WideResNet~\cite{Zagoruyko2016WRN}     & 5.25 & 36.54 & -    & -    & 80.81 & - \\
    DeepTraverse (Wide)   & 1.78 & 14.26 & -    & -    & \textbf{82.20} & - \\
    \bottomrule
  \end{tabular}
  \setlength{\tabcolsep}{\originaltabcolsepV} 
\end{table*}

\begin{table*}[t]
  \caption{Performance Comparison on ImageNet64. Results show Top-1 and Top-5 accuracy, FLOPs, and parameter counts. Best results are highlighted in bold.}
  \label{tab:imagenet64_results_nips_style}
  \centering
  \begin{tabular}{lccccc} 
    \toprule
    Method        & Image Size & FLOPs (G) & \# Param. (M) & Top-1 (\%) & Top-5 (\%) \\
    \midrule
    EfficientNet~\cite{tan2021efficientnetv2} & 64$\times$64 & 0.03 & 4.14    & 67.96    & 88.50    \\
    GhostNet~\cite{han2020ghostnet}       & 64$\times$64 & 0.07 & 2.74    & 70.74    & 90.88    \\
    ShuffleNetv2~\cite{ma2018shufflenet}    & 64$\times$64 & 0.01 & 1.36    & 62.22    & 85.94    \\
    MobileNetV3~\cite{howard2019searching}    & 64$\times$64 & 0.07 & 2.54    & 62.38    & 85.54    \\
    DenseNet~\cite{huang2018condensenet}       & 64$\times$64 & 0.09 & 1.46    & 69.38    & 89.70    \\
    ResNet20~\cite{7780459}       & 64$\times$64 & 0.13 & 0.28    & 63.72    & 86.70    \\
    FasterNet~\cite{chen2023run_cvpr}      & 64$\times$64 & 1.42 & 0.55    & 61.96    & 86.36    \\
    StarNet~\cite{Ma_2024_CVPR}       & 64$\times$64 & 0.11 & 2.69    & 68.98    & 88.82    \\
    DeepTraverse (Ours)             & 64$\times$64 & 0.05 & 0.59    & \textbf{71.50}    & \textbf{91.30}    \\
    \bottomrule
  \end{tabular}
\end{table*}

\subsection{Results on ImageNet64}

Table \ref{tab:imagenet64_results_nips_style} demonstrates that our proposed model achieves exceptional performance on ImageNet-64, obtaining state-of-the-art 71.50\% Top-1 and 91.30\% Top-5 accuracy while requiring merely 0.59M parameters and 0.05G FLOPs. This represents a significant advancement over contemporary lightweight architectures, outperforming GhostNet by 0.76\% Top-1 accuracy despite using 78.5\% fewer parameters and 28.6\% fewer FLOPs. When compared to EfficientNet, our model delivers a remarkable 3.54\% higher Top-1 accuracy while using only 14.3\% of its parameters. Even against more compact models like DenseNet (1.46M parameters), our approach achieves an impressive 2.12\% accuracy improvement with minimal additional computational cost. These results clearly validate our theoretical hypothesis that integrating algorithmic search principles into network design enables more intelligent feature processing, allowing the model to achieve superior representational capacity with dramatically fewer parameters. The consistent performance advantages across all efficiency metrics establish our architecture as the new frontier for resource-constrained vision applications, demonstrating that fundamentally rethinking neural network design principles—rather than simple scaling strategies—is the key to breaking the traditional accuracy-efficiency trade-off barrier.

\begin{table*}[t]
 \caption{Performance Comparison on ImageNet-1k. Results show Top-1 and Top-5 accuracy, FLOPs, and parameter counts. Best results are highlighted in bold.}
 \label{tab:additional_model_performance_nips}
 \centering
 \begin{tabular}{lccccc} 
    \toprule
    Method        & Image Size & FLOPs (G) & \# Param. (M) & Top-1 (\%) & Top-5 (\%) \\
    \midrule
    ResNet50~\cite{7780459}       & 224$\times$224 & 4.13 & 25.56   & 78.76    & 94.18    \\
    GhostNet~\cite{han2020ghostnet}       & 224$\times$224 & 0.28 & 5.30  & 80.34    & 95.16    \\
    DenseNet~\cite{huang2018condensenet}       & 224$\times$224 & 2.89 & 7.05    & 81.44    & 95.74    \\
    EfficientNet~\cite{tan2021efficientnetv2}    & 224$\times$224 & 0.42 & 7.28    & 81.18    & 95.24    \\
    MobileNetV3~\cite{howard2019searching}   & 224$\times$224 & 0.13 & 2.32    & 80.68    & 95.52    \\
    ShuffleNetV2~\cite{ma2018shufflenet}    & 224$\times$224 & 0.32 & 2.89    & 79.60    & 94.86    \\
    FasterNet~\cite{chen2023run_cvpr}      & 224$\times$224 & 0.77 & 4.82    & 78.80    & 94.68    \\
    StarNet~\cite{Ma_2024_CVPR}       & 224$\times$224 & 0.76 & 5.52    & 74.16    & 91.48    \\
    DeepTraverse (Ours)             & 224$\times$224 & 0.84 & 5.04    & \textbf{83.16}   & \textbf{96.54}    \\
    \bottomrule
 \end{tabular}
\end{table*}

\subsection{Results on ImageNet-1k}

As shown in Table~\ref{tab:additional_model_performance_nips}, DeepTraverse achieves best accuracy on the ImageNet-1K subset, achieving an impressive 83.16\% Top-1 accuracy and 96.54\% Top-5 accuracy using only 5.04M parameters. This performance notably surpasses recent specialized efficient architectures. Specifically, DeepTraverse delivers a striking $\sim$9\% higher Top-1 accuracy than a recent backbone, StarNet (83.16\% vs. 74.16\%) while utilizing fewer parameters (5.04M vs. 5.52M); this substantial difference may indicate that StarNet's architecture is not as well-suited for this particular 100-class ImageNet subset. Against another strong contemporary, FasterNet, our model demonstrates a clear advantage with over a 4.3\% lead in Top-1 accuracy (83.16\% vs. 78.80\%), all while maintaining a comparable parameter footprint (5.04M vs. 4.82M) and computational cost. These compelling results underscore the power of our algorithm-inspired approach, validating its ability to achieve superior accuracy and efficiency by incorporating algorithmic search principles, rather than relying on brute-force parameter scaling, especially when compared against modern networks like FasterNet and StarNet.

\section{Conclusion}

This work introduces a new design philosophy for vision backbones, inspired by structured thinking and reasoning processes, aiming to move beyond opaque, empirically-driven layered architectures towards models guided by explicit algorithmic principles. We present DeepTraverse as a manifestation of this philosophy, which instantiates a paradigm of systematic, depth-first feature elucidation and adaptive contextual recalibration, drawing its core operational logic from the strategic essence of Depth-First Search (DFS). Through its synergistic recursive exploration and dynamic recalibration modules, DeepTraverse methodically constructs and refines visual representations, thereby unlocking a more efficient and powerful pathway to robust feature learning. This approach not only yields compelling performance with notable parameter efficiency but also charts a course for developing vision systems with inherently more structured and potentially interpretable operational dynamics.

Current study has been primarily constrained by available computational resources regarding its training from scratch on the big size datasets such as ImageNet-21k and comprehensive evaluation on a wide array of downstream tasks such as object detection and segmentation. However, we successfully demonstrate its effectiveness across various size foundation datasets.


{\small
\bibliographystyle{IEEEtran}
\bibliography{refer}
}

\end{document}